# Performance Characterization of Image Feature Detectors in Relation to the Scene Content Utilizing a Large Image Database

Bruno Ferrarini[1], Shoaib Ehsan[1], Naveed Ur Rehman[2] and Klaus D. McDonald-Maier[1]

[1]School of Computer Science and Electronic Engineering, University of Essex, Wivenhoe Park, Colchester, UK
[2]Department of Electrical Engineering, COMSATS Institute of Information Technology, Islamabad, Pakistan
bferra@essex.ac.uk

*Abstract* - Selecting the most suitable local invariant feature detector for a particular application has rendered the task of evaluating feature detectors a critical issue in vision research. No state-of-the-art image feature detector works satisfactorily under all types of image transformations. Although the literature offers a variety of comparison works focusing on performance evaluation of image feature detectors under several types of image transformation, the influence of the scene content on the performance of local feature detectors has received little attention so far. This paper aims to bridge this gap with a new framework for determining the type of scenes, which maximize and minimize the performance of detectors in terms of repeatability rate. Several state-of-the-art feature detectors have been assessed utilizing a large database of 12936 images generated by applying uniform light and blur changes to 539 scenes captured from the real world. The results obtained provide new insights into the behaviour of feature detectors.

*Keywords* - Local Feature Detection; Evaluation Framework; Performance Analysis

## I. INTRODUCTION

Local feature detection has been a challenging problem for the computer vision community for a long time. A large number of different approaches have been proposed so far, thus turning evaluation of image feature detectors into an active research topic in the last decade or so. Most of the evaluations available in the literature focus mainly on characterizing feature detectors' performance under different image transformations without analysing in detail the effects of the scene content. In [1], the feature tracking capabilities of some corner detectors are assessed utilizing static image sequences of a few different scenes. Although the results permit us to infer a dependency of the detectors' performance on the scene content, the methodology followed is not intended to highlight and formalize such a relationship as no classification is assigned to the scenes. The comparison work in [2] gives a formal definition for textured and structured scenes and shows the repeatability rates of six feature detectors. The results provided by [2] show that the content of the scenes influences the repeatability but the framework utilized and the small number of scenes included in the datasets employed [3] do not provide a comprehensive insight into the behaviour of the feature detectors with different types of scenes. In [4], the scenes are classified by the complexity of their 3D structures in complex and planar categories. The repeatability results reveal how detectors perform for those two categories. The limit in the generality of the analysis done in [4] is due to the small number and limited variety of the scenes employed, whose content are mostly human-made.

This paper proposes a new approach to estimate the effect of the scene content on the performance of local image feature detectors by introducing a metric to measure their bias towards a scene with particular characteristics. The methodology proposed utilizes the improved repeatability criterion presented in [5] as a measure of the performance of feature detectors, and the large database [6] of images consisting of 539 different real-world scenes containing a wide variety of different elements. The reminder of the paper is organized as follows. Section II provides an overview of the related work in the field of feature detector evaluation and scene taxonomy. In Section III, the proposed evaluation framework is introduced. Section IV describes the image database employed. In Section V, the results obtained utilizing the proposed framework are presented and discussed. Finally, the conclusions are given in Section VI.

## II. RELATED WORK

The contributions to the evaluation of local feature detectors are numerous and vary based on: the metric used for quantifying the detector performance, the methodology followed and the image database used. Repeatability is a desirable property for feature detectors as it measures the grade of independence of the feature detector from changes in the imaging conditions. For this reason, it is frequently used as a measure of performance of local feature detectors. A definition of repeatability is given in [7] where, together with the information content, it is utilized as a metric for comparing six feature detectors. A refinement of the definition of repeatability is given in [8], and used for assessing six state-of-the-art feature detectors in [2] under several types of transformations on textured and structured scenes. Two criteria for an improved repeatability measure are introduced in [5] that provide results which are more consistent with the actual performance of several popular feature detectors on the widely-used Oxford datasets [3]. Moreover, repeatability is used as a metric for performance evaluation in [9] and [4] that utilize non-planar, complex and simple scenes. The performance of feature detectors has also been assessed employing metrics other than repeatability. The performance measure in [10] is completeness, while feature coverage is used as a metric in [11]. The feature detectors have also been evaluated in the context of a specific application, such as in [1], where corner

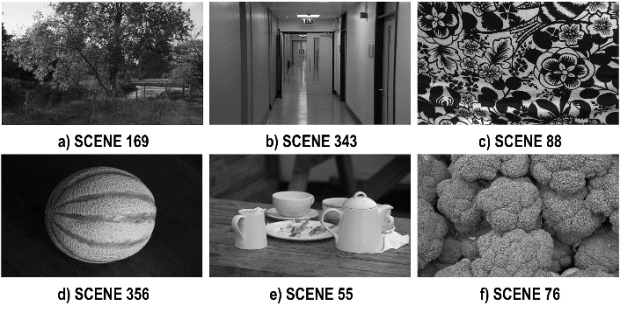

Figure 1. The images (a-f) show a sample of the scenes in the database used for the experiments.

detectors are assessed in the context of point feature tracking applications.

## III. THE PROPOSED EVALUATION FRAMEWORK

The proposed framework has been designed by keeping in mind the objective of evaluating the influence of scene content on the performance of a wide variety of state-of-the-art feature detectors. A proper application of such a framework requires a large image database ($I$) organized in a series of $n$ datasets. Each dataset needs to contain images from a single scene with different amounts of an image transformation. The images included in such a database should be taken from a large variety of different real-world scenarios. The proposed framework consists of the steps discussed below.

### A. Repeatability data

The repeatability rates required by the proposed framework are computed utilizing the criterion described in [5], whose consistency with the actual performance of a wide variety of feature detectors has been proved across well-established datasets [3]. As proposed in [5], the repeatability rate is defined as $Repeatability = N_{rep}/N_{ref}$, where $N_{rep}$ is the total number of repeated features and $N_{ref}$ is the number of interest points in the common part of the reference image. Let $A$ and $P$ be the sets of indices representing the $m$ discrete amounts of transformation and the scenes respectively.

$$A = \{1,2,3,\ldots\ldots,m\} \quad (1)$$

$$P = \{1,2,3,\ldots\ldots,n\} \quad (2)$$

where $m$ corresponds to the maximum amount of transformation and 1 relates to the reference image (zero transformation); $n$ is the total number of scenes and each scene is utilized to build one separate dataset, thus finally resulting in $n$ datasets in total. Let $B_{kd}$ be the set of repeatability rates computed for step $k \in A$ (corresponding to $k^{th}$ image transformation amount) for a feature detector $d$ across $n$ datasets (which implies repeatability values for $n$ scenes):

$$B_{kd} = \{B_{1kd}, B_{2kd}, \ldots\ldots, B_{nkd}\} \quad (3)$$

Each set $B_{kd}$ contains $n$ repeatability ratios, one for each dataset.

### B. Scene Rankings

The top and lowest rankings for each detector $d$ are built selecting the $j$ highest and the lowest repeatability scores at the amount $k$ of image transformation. Let $T_{kd}(j)$ and $W_{kd}(j)$ the sets containing the indices of the scenes whose repeatability falls in the top and lowest ranking respectively:

$$T_{kd}(j) = \{S_{kd(1)}, S_{kd(2)}, \ldots\ldots, S_{kd(i)}\} \quad (4)$$

$$W_{kd}(j) = \{S_{kd(n)}, S_{kd(n-1)}, \ldots\ldots, S_{kd(n-i+1)}\} \quad (5)$$

where $S_{kd(i)} \in P$ is the scene index corresponding to the $i^{th}$ highest repeatability score obtained by the detector $d$ for the scene under the amount $k$ of transformation. Thus, accordingly with this notation, $S_{kd(1)}$ is the scene for which the detector scored the best repeatability score, $S_{kd(2)}$ corresponds to the second highest repeatability rate, $S_{kd(3)}$ to the third highest and so on, until $S_{kd(n)}$ which is for the lowest one.

### C. Scene classification

The scenes are attributed with three labels on the basis of human judgement. As described in Table 1, each label is dedicated to a particular property of the scene and has been assigned independently from the others. These attributes are: the location type ($f$) which may take the label outdoor or indoor, the type of the elements contained ($g$) which may take the label natural or human-made, and the perceived complexity of the scene ($h$) which may take the label simple or complex. Figure 1 shows a sample of the scenes from the image database [6] utilized for the experiments. Scene 169 is tagged as outdoor and, along with scene 76 and 356, contains natural elements. The scenes 343, 88 and 55 are labelled as human-made and the first one is also classified as indoor. The scene 88 is categorized as a simple scene as it includes a few edges delimiting well contrasted areas. Although main structures (broccolis' borders) can be identified in scene 76, the rough surface of the broccolis is information rich so results in labelling this scene as complex.

Figure 2. Trait indices expressed in percentage for uniform light reduction at three amounts of transformation. For both the top (green, left half) and lowest (red, right half) twenty rankings are shown the outdoor (*F*), human-made (*G*) and simple (*H*) indices.

| Light Reduct. | | TOP 20 | | | LOWEST 20 | | | Light Reduct. | | TOP 20 | | | LOWEST 20 | | |
|---|---|---|---|---|---|---|---|---|---|---|---|---|---|---|---|
| | | F | G | H | F | G | H | | | F | G | H | F | G | H |
| EBR | 10% | 50 | 85 | 70 | 45 | 20 | 20 | IBR | 10% | 30 | 70 | 65 | 65 | 40 | 25 |
| | 40% | 45 | 90 | 65 | 55 | 20 | 30 | | 40% | 35 | 50 | 25 | 75 | 20 | 25 |
| | 60% | 60 | 90 | 60 | 35 | 25 | 35 | | 60% | 40 | 55 | 30 | 50 | 30 | 30 |
| HARAFF | 10% | 45 | 85 | 85 | 15 | 35 | 20 | MSER | 10% | 25 | 95 | 80 | 65 | 50 | 45 |
| | 40% | 50 | 80 | 85 | 20 | 50 | 35 | | 40% | 20 | 100 | 85 | 60 | 50 | 55 |
| | 60% | 45 | 85 | 85 | 35 | 45 | 50 | | 60% | 30 | 100 | 85 | 55 | 60 | 50 |
| HARLAP | 10% | 45 | 85 | 85 | 15 | 35 | 20 | SALIENT | 10% | 55 | 35 | 10 | 35 | 25 | 85 |
| | 40% | 50 | 80 | 85 | 20 | 50 | 35 | | 40% | 30 | 50 | 15 | 30 | 40 | 85 |
| | 60% | 45 | 85 | 85 | 35 | 45 | 50 | | 60% | 25 | 35 | 10 | 45 | 45 | 65 |
| HESAFF | 10% | 55 | 90 | 70 | 20 | 30 | 15 | SFOP | 10% | 30 | 35 | 20 | 50 | 45 | 50 |
| | 40% | 50 | 100 | 75 | 30 | 40 | 30 | | 40% | 35 | 35 | 20 | 60 | 45 | 35 |
| | 60% | 50 | 95 | 75 | 25 | 55 | 50 | | 60% | 30 | 40 | 30 | 55 | 40 | 30 |
| HESLAP | 10% | 50 | 90 | 70 | 15 | 40 | 30 | SURF | 10% | 60 | 65 | 55 | 0 | 50 | 40 |
| | 40% | 45 | 90 | 75 | 25 | 45 | 30 | | 40% | 55 | 70 | 50 | 10 | 45 | 30 |
| | 60% | 50 | 90 | 75 | 25 | 55 | 50 | | 60% | 50 | 80 | 75 | 35 | 35 | 35 |

Table 1 Classification labels and criteria.

| | | |
|---|---|---|
| Location Type | non-Outdoor | Indoor scene and close-up of a single or of a few objects. |
| | Outdoor | The complement of above. |
| Object Type | Human-made | Elements are mostly artificial. |
| | Natural | Elements are mostly natural. |
| Complexity | Simple | A large number of edges with fractal-like shapes. |
| | Complex | A large number of edges with fractal-like shapes. |

*D. Ranking traits indices*

The labels of the scenes included in the rankings, (4) and (5), are examined in order to determine the dominant types of scenes. For each ranking $T_{kd}(j)$ and $W_{kd}(j)$, the ratios of scenes classified as outdoor, human-made and simple are computed. Thus, three ratios are associated with each ranking where higher values mean a higher share of the scene type associated:

$$\forall S_i \in T_{kd}: T_{kd}.[F,G,H] = \frac{\sum S_i.[f,g,h]}{j} \quad (6)$$

$$\forall S_i \in W_{kd}: W_{kd}.[F,G,H] = \frac{\sum S_i.[f,g,h]}{j} \quad (7)$$

These vectors contain three measures, which represent the extent of the bias of detectors. For example, if a top ranking presents $F = 0.1$, $G = 0.25$ and $H = 0.8$, it can be concluded that the detector, for the given amount of image transformation, works better with scenes whose elements are mostly natural (low $G$), with simple edges (high $H$) and that are not outdoor (low $F$). As opposed to that, if the same indices were for a worst ranking, it could be concluded that the detector obtains its worst results for non-outdoor ($F$) and natural ($G$) scenes with low edge complexity ($H$).

IV. THE IMAGE DATABASE

The image database used for the experiment is discussed in this section and is available at [6]. It contains a large number of images from real-world scenes which are representative of a wide variety of different environment and natural elements. Each of the datasets belonging to the database, includes a reference image of a single scene and several images generated by applying a transformation at several discrete steps of amounts which are 9 for Gaussian blur and 13 for uniform light change. Thus, each scene has been used for generating two datasets, one for each of the transformations utilized, for a total of 539 x 2 datasets. Several well-established datasets, such as [3], are available for evaluating local feature detectors but are not suitable for use with the proposed framework due to the relatively small number and less variety of scenes included. Among the Oxford datasets [3], Leuven offers images under uniform light changes but the number of images in that dataset is only six. Although the database employed in [12] for assessing several feature detectors under different light conditions contains a large number of images, the number of scenes are limited to 60. Moreover, these scenes were captured in a highly controlled environment and do not capture real-world scenarios completely. The images included in the database utilized for this work have a resolution of 717 × 1080 pixels and consist of 539 real-world scenes. Each transformation is applied in several discrete steps to each of the scenes. The Gaussian blur amount is varied in 10 discrete steps from 0 to 4.5σ while the amount of light is reduced from 100% to 10%. Thus, the database includes a dataset of 10 or 14 images for each of the 539 scenes and transformation type for a total of 12936 images.

| Blur | | TOP 20 | | | LOWEST20 | | | Blur | | TOP 20 | | | LOWEST 20 | | |
|---|---|---|---|---|---|---|---|---|---|---|---|---|---|---|---|
| | | F | G | H | F | G | H | | | F | G | H | F | G | H |
| EBR | 0.5 | 50 | 85 | 70 | 70 | 10 | 10 | IBR | 0.5 | 15 | 60 | 90 | 75 | 40 | 20 |
| | 2.0 | 30 | 95 | 75 | 55 | 35 | 15 | | 2.0 | 20 | 70 | 95 | 75 | 50 | 30 |
| | 3.0 | 30 | 85 | 80 | 50 | 30 | 30 | | 3.0 | 15 | 80 | 95 | 65 | 40 | 25 |
| HARAFF | 0.5 | 10 | 35 | 80 | 40 | 45 | 25 | MSER | 0.5 | 20 | 90 | 90 | 45 | 45 | 40 |
| | 2.0 | 10 | 45 | 95 | 35 | 55 | 25 | | 2.0 | 30 | 95 | 95 | 25 | 35 | 45 |
| | 3.0 | 20 | 25 | 95 | 40 | 45 | 20 | | 3.0 | 35 | 95 | 100 | 45 | 30 | 40 |
| HARLAP | 0.5 | 20 | 35 | 80 | 35 | 50 | 30 | SALIENT | 0.5 | 10 | 85 | 75 | 80 | 30 | 20 |
| | 2.0 | 10 | 45 | 95 | 35 | 55 | 25 | | 2.0 | 10 | 50 | 75 | 80 | 15 | 0 |
| | 3.0 | 20 | 30 | 95 | 40 | 40 | 20 | | 3.0 | 5 | 35 | 85 | 90 | 25 | 5 |
| HESAFF | 0.5 | 30 | 70 | 75 | 30 | 35 | 25 | SFOP | 0.5 | 10 | 80 | 80 | 80 | 65 | 25 |
| | 2.0 | 25 | 60 | 100 | 50 | 35 | 15 | | 2.0 | 15 | 60 | 80 | 70 | 50 | 20 |
| | 3.0 | 25 | 50 | 100 | 45 | 35 | 15 | | 3.0 | 20 | 55 | 100 | 70 | 50 | 20 |
| HESLAP | 0.5 | 45 | 80 | 65 | 30 | 35 | 25 | SURF | 0.5 | 20 | 30 | 85 | 60 | 55 | 35 |
| | 2.0 | 20 | 65 | 95 | 45 | 40 | 20 | | 2.0 | 20 | 45 | 90 | 25 | 50 | 25 |
| | 3.0 | 15 | 50 | 100 | 40 | 35 | 15 | | 3.0 | 20 | 40 | 95 | 30 | 45 | 20 |

Figure 3. Trait indices expressed in for Gaussian blur at three amounts of transformation ($\sigma$). For both the top (green, left half) and lowest (red, right half) twenty rankings are shown the outdoor ($F$), human-made ($G$) and simple ($H$) indices.

V. RESULTS

The proposed framework has been applied for producing the top and bottom rankings for a set of several feature detectors which are representative of a wide variety of different approaches [13] and includes the following: Edge-Based Region (EBR) [14], Harris-Affine (HARAFF), Hessian-Affine (HESAFF) [15], Maximally Stable External Region (MSER) [16], Harris-Laplace (HARLAP), Hessian-Laplace (HESLAP) [8], Intensity-Based Region (IBR) [17], SALIENT [18], Scale-invariant Feature Operator (SFOP) [19] and Speeded Up Robust Feature (SURF) [20]. The repeatability data is obtained for each transformation type utilizing the image database discussed in Section IV. This data is collected using the authors' original programs with control parameter values suggested by them. The feature detector parameters could be varied in order to obtain a similar number of extracted features for each detector. However, this has a negative impact on the repeatability of a detector [8] and is therefore not desirable for such an evaluation. The top and lowest rankings for various transformation amounts are built by considering the top and the lowest twenty scenes ($j = 20$) respectively and finally their trait indices are computed. Due to space constraints, only a sample of the results obtained is shown by selecting three particular amounts of image transformation for each considered type. Figure 2 is dedicated to the traits indices obtained from the uniform light change datasets while Figure 3 shows results from Gaussian blur datasets.

*A. Uniform light reduction*

Under uniform light reduction, most state-of-the-art feature detectors achieve their highest repeatability rates with scenes labelled as simple and human-made. Some good examples are EBR, HARAFF, HARLAP, HESAFF, HESLAP and MSER that show such a bias. Indeed, they are characterized by high values of $G$ and $H$ ($\geq 70\%$) on the top twenty scenes and much lower values on the lowest twenty ($\leq 50\%$). An exception to this trend is represented by SFOP and SALIENT whose bias is towards natural and complex scenes. Only IBR, MSER, SFOP and SALIENT manifest an explicit preference for non-outdoor scenes. The other detectors perform worst on non-outdoor scenes as can be concluded by observing the values of *F* for the lowest twenty scenes, which are frequently below 40%. Finally, the indices vary little with light reduction amount increasing. Indeed, not in a single case, the scene type preference of any detector is inverted along the range of the amount variation.

*B. Blur changes*

Under Gaussian blurring, the top twenty scenes are those containing artificial elements and simple edges. Indeed, all the detectors present very high values for *H* (in most cases 80% or above) in the top twenty rankings and very low in the lowest rankings (mostly below 40%). IBR and MSER show an inclination towards non-indoor scenes and artificial objects. SALIENT and SFOP achieve a particularly high share (70% or above) of outdoor images in their lowest rankings at each amount of transformation. From the observation of the data, a direct relationship between *H* value and the blur amount in the top rankings can be inferred. Indeed, Gaussian blur is a type of transformation which soften the edges by smoothening the higher frequencies component of images. Thus, the edges become more difficult to detect and those belonging to the smallest details of images may be masked completely with a consequent reduction of repeatability rates.

*C. Discussion*

The results show that the biases of some detectors are affected by the type and the amount of transformation. For example, the biases of SALIENT, the only entropy-based detector [18], resulted particularly variable with the transformation type. On the other hand, MSER [16] which is a segmentation-based detector, presented the most stable scene type share in both groups of rankings, top and lowest, across all the types and amounts of transformation.

## VI. CONCLUSIONS

For several state-of-the-art feature detectors, the dependency of the repeatability from the input scene type has been investigated utilizing a large database composed of images from a wide variety of human-classified scenes under uniform light reduction and Gaussian blur transformations. Although the utilized human-based classification method includes just three independently assigned labels, it is enough to prove that the feature detectors tend to score their highest and lowest repeatability scores with a particular type of scene and how such a tendency is strong. The proposed framework is intended to help visual system designers in maximizing the performance of the applications which utilize local image feature detectors. Indeed, the framework can be employed for identifying the detectors that perform better with the type of scene most common in an application before any further task-oriented evaluation (e.g. [1]) which, at that point, it would be carried out on a smaller set of local feature detectors. For example, for an application which deals mainly with indoor scenes, the detectors should be short-listed are MSER, IBR and SFOP that have been proven to achieve their highest repeatability rates with non-outdoor scenes under both light reduction and blurring. On the other hand, if an application is intended for working in an outdoor environment, EBR should be one of the considered local feature detectors, especially under light reduction. In brief, the framework proposed permits us to characterize the feature detector by scene content and, at the same time, represents a useful tool for facilitating the design of those visual applications, which utilize a local feature detector stage.